\documentclass{article}
\usepackage{spconf,amsmath,graphicx}
\usepackage{amssymb}

\usepackage[style=ieee, maxnames=5, minnames=1, giveninits=true, backend=biber, citestyle=numeric-comp]{biblatex}
\usepackage{enumitem}
\setlist{nosep, leftmargin=14pt}

\usepackage{mwe} 
\usepackage{multirow}

\title{Understanding the Transfer Limits of Vision Foundation Models}
\name{
\begin{tabular}{c}
     Shiqi Huang, Yipei Wang, Natasha Thorley, Alexander Ng, Shaheer Saeed, Mark Emberton,\\Shonit Punwani, Veeru Kasivisvanathan, Dean Barratt, Daniel Alexander, Yipeng Hu
\end{tabular}}
\address{University College London}
\begin{document}
%
\maketitle
\begin{abstract}
Foundation models leverage large-scale pretraining to capture extensive knowledge, demonstrating generalization in a wide range of language tasks. By comparison, vision foundation models (VFMs) often exhibit uneven improvements across downstream tasks, despite substantial computational investment. We postulate that this limitation arises from a mismatch between pretraining objectives and the demands of downstream vision-and-imaging tasks. Pretraining strategies like masked image reconstruction or contrastive learning shape representations for tasks such as recovery of generic visual patterns or global semantic structures, which may not align with the task-specific requirements of downstream applications including segmentation, classification, or image synthesis. To investigate this in a concrete real-world clinical area, we assess two VFMs, a reconstruction-focused MAE-based model (ProFound) and a contrastive-learning-based model (ProViCNet), on five prostate multiparametric MR imaging tasks, examining how such task alignment influences transfer performance, i.e., from pretraining to fine-tuning. Our findings indicate that better alignment between pretraining and downstream tasks, measured by simple divergence metrics such as maximum-mean-discrepancy (MMD) between the same features before and after fine-tuning, correlates with greater performance improvements and faster convergence, emphasizing the importance of designing and analyzing pretraining objectives with downstream applicability in mind.
\end{abstract}
\begin{figure}
    \centering
    \includegraphics[width=\linewidth]{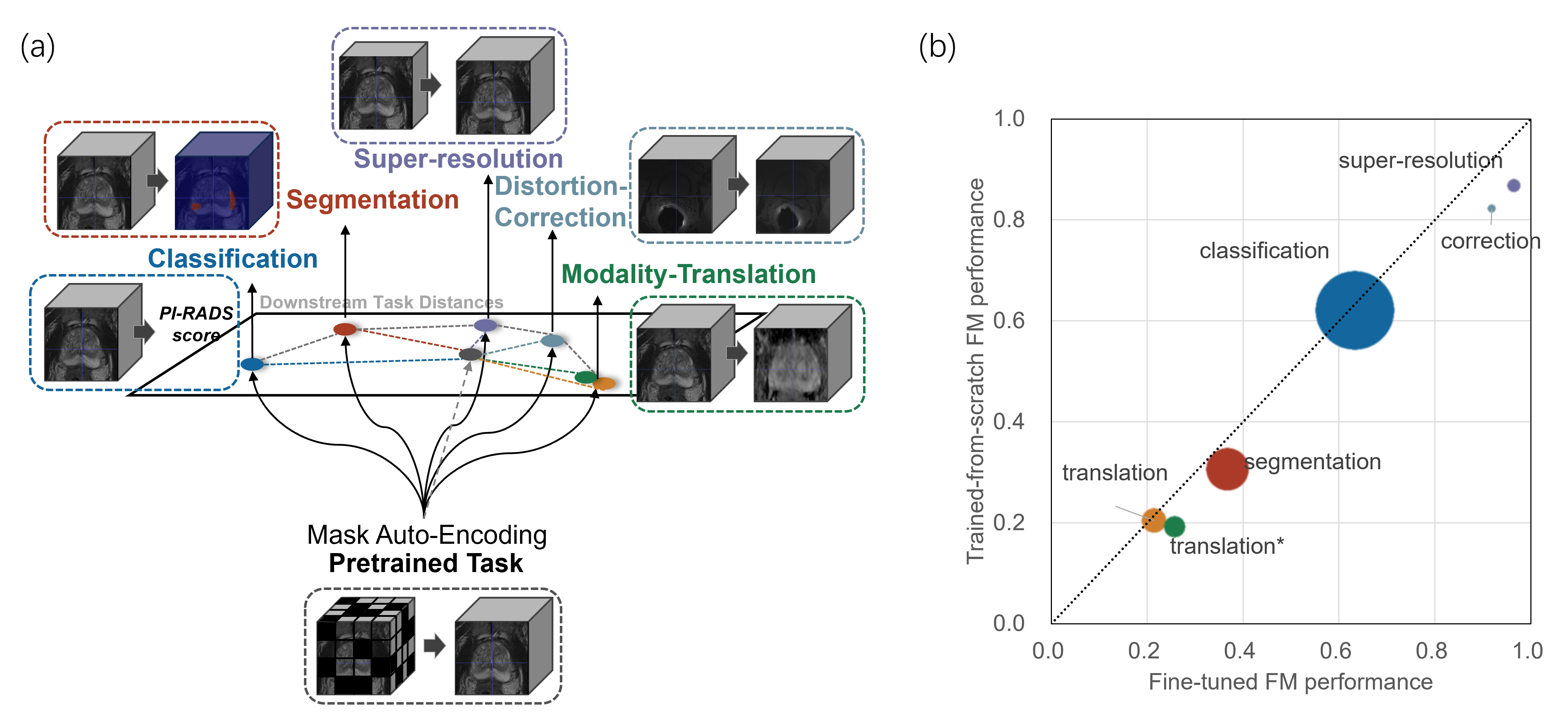}
    \caption{a) Illustration of the pretraining-downstream task alignment using ProFound. b) Scatter plot showing ProFound performance trained from scratch (x-axis) versus fine-tuned from pretrained weights (y-axis). Color encodes task identity, and marker size reflects D2S value. }
    \label{fig:motivation}
    \vspace{-1.5em}
\end{figure}

\section{Introduction}
\label{sec:intro}



Language foundation models achieve useful generalization largely because their pretraining and downstream tasks share a clear conceptual alignment - both revolve around predicting the next word token. This tight correspondence allows the pretraining objective to directly benefit multiple downstream tasks such as translation, summarization, and reasoning~\cite{achiam2023gpt,touvron2023llama}.

By contrast, for current vision foundation models (VFMs), the relationship between pretraining objectives and downstream tasks is much less straightforward~\cite{bi2024learning,chen2024internvl,moor2023foundation,awais2025foundation}. It remains unclear to what extent common pretraining paradigms - such as masked image reconstruction~\cite{he2022mae,caro2023brainlm,jiang2024m4oe} or contrastive learning~\cite{oquab2023dinov2,lee2025prostate,suo2025automatic} - are aligned with practical applications like segmentation
. For instance, masked image reconstruction may share resemblance to super-resolution tasks, whereas contrastive objectives that discriminate between spatial augmentations might be partially analogous to classification or segmentation tasks that depend on morphological differences.

However, such similarity between tasks - task alignment - has yet been quantified and arguably under-studied in real-world clinical imaging applications. This misalignment, to varying degrees, between pretraining and downstream tasks, may have contributed to the mixed benefits of VFMs reported in recent literature~\cite{zamir2018taskonomy,gontijo2020tradeoffs,cekmeceli2024vision}.
Moreover, the data and computational requirements for different VFMs are still poorly characterized, further complicating fair comparison.

In this work, we investigate these issues in a specific prostate imaging context of our interest by evaluating two VFMs: ProFound, a reconstruction-oriented MAE-based model, and ProViCNet~\cite{lee2025prostate}, a contrastive-learning-based model-across five prostate multiparametric MRI tasks: classification, segmentation, super-resolution, distortion correction, and modality translation. We analyze how the degree of task alignment influences transfer performance during fine-tuning, focusing particularly on data and compute efficiency.

Contributions of this work are summarized:
1) A task-driven perspective attributing the limited transfer gains of VFMs to intrinsic discrepancies between pretraining and downstream objectives;
2) A quantitative method to measure pretraining-downstream alignment via distributional similarity;
3) The first set of quantitative evidence to demonstrate that task alignment relates to transfer performance across pretraining paradigms and downstream tasks.


\section{Pretraining and fine-tuning}
\label{sec:framework}
\subsection{Pretraining}
Recent prostate-specific vision foundation models have been developed with large-scale prostate cancer imaging data.
\textbf{ProFound}~\footnote{https://github.com/pipiwang/ProFound.git} is a vision foundation model pretrained on $\sim$5,000 prostate-MRI exams (T2w, ADC, high-b DWI) from PI-CAI~\cite{saha2024picai} and several other private trial datasets. It uses a ConvNeXt V2 Tiny backbone trained via \textit{masked autoencoding (MAE) objective} to learn generalizable imaging features.
\textbf{ProViCNet}~\footnote{https://github.com/pimed/ProViCNet.git} is a multi-modal foundation model pretrained on 4,401 patient cases collected from six institutions~\cite{lee2025prostate}, using multiparametric MRI and some TRUS images. The model employs a 3D-enhanced vision transformer backbone pretrained on DINOv2 with patch-level \textit{contrastive learning} guided by biopsy-confirmed lesion annotations.


\subsection{Fine-tuning}

ProFound and ProViCNet are evaluated across five downstream prostate MRI tasks: \textit{classification}, \textit{segmentation}, \textit{super-resolution}, \textit{distortion correction} and \textit{modality translation} through task-specific fine-tuning strategies. 

\textbf{ProFound} follows the lightweight fine-tuning protocol in MAE~\cite{he2022mae}. The pretrained encoder is frozen, while only the task-specific decoder or head - when required - is optimized using limited labeled data. For classification and segmentation, we follow the settings provided in the repo. For super-resolution, a subpixel head is added to adjust channel dimensions, while distortion correction and modality translation are formulated the same as fine-tuning segmentation tasks with softmax outputs for normalized intensities.

\textbf{ProViCNet} adopts a minimal fine-tuning strategy following DINOv2~\cite{oquab2023dinov2}. Specifically, the last two encoder blocks and normalization layers are unfrozen, while the task-specific head (if required) is jointly optimized. As ProViCNet was originally designed for segmentation, the fine-tuning setups for segmentation, distortion correction, and modality translation follow its original implementation with softmax outputs. For classification, a three-layer MLP classifier is appended, and for super-resolution, a subpixel head is used.

For both models, the optimizer is AdamW with an initial learning rate of 1e-4 and a cosine annealing over 100 epochs. Task-specific loss functions for each downstream task are described in Sec.\ref{sec: downstream}.


\section{Experiments}
\label{sec:exp}

\subsection{Downstream Tasks}
\label{sec: downstream}
For downstream fine-tuning, we benchmark foundation model (FM) and several baselines, for individual tasks:  
(1) the pretrained foundation model fine-tuned on downstream task (fine-tuned FM); (2) the pretrained foundation model without task-specific fine-tuning (pretrained FM); (3) an architectural counterpart with identical structure with random weights (random-weight FM); (4) an architectural counterpart with identical structure trained from scratch on downstream task (trained-from-scratch FM); and (5) specialized task-specific models developed and trained specifically for each downstream task (e.g., unet). The trained-from-scratch FM, fine-tuned FM and specialized task-specific models are each trained/fine-tuned on the same task-specific datasets using an identical loss. Performance on each downstream task is evaluated on the same test set with the same metric.


\noindent\textbf{1. PI-RADS Classification} predicts the PI-RADS score of each MR image, formulated as $f_{cls}: \mathbb{R}^{H\times W\times D} \rightarrow \{1,2,...,K\}$ with $K=4$. We utilize the PROMIS dataset~\cite{ahmed2017diagnostic}, which contains 740 patient cases with corresponding multiparametric MRI (mpMRI), clinically assigned PI-RADS scores, and lesion segmentation masks. For this task, T2-weighted (T2W) modality is used as input~\cite{yi2025t2}, with each 3D volume resampled to a spatial size of $200 \times 200 \times 96$. The dataset is split into training, validation, and test sets with 555, 37, and 148 samples, respectively. To mitigate class imbalance, we employ a weighted cross-entropy loss. Model performance is evaluated using specificity at a fixed sensitivity of 0.9. Resnet50~\cite{he2016resnet} is employed as specialized task-specific model for PI-RADS classification task.

\noindent\textbf{2. Lesion Segmentation} is defined as $ f_{\text{seg}}: \mathbb{R}^{H \times W \times D} \rightarrow \{1, 2, \dots, L\}^{H \times W \times D} $, where $L$ is the number of lesion classes~\cite{duran2022prostattention}. We again employ the PROMIS dataset~\cite{ahmed2017diagnostic}, where the T2W modality and a randomly selected diffusion modality (either ADC or high-b) serve as inputs, with the corresponding prostate cancer lesion masks as ground truth. Each volume is resampled to $200 \times 200 \times 96$, with identical data splits as in the classification task. Fine-tuning (or training) uses the Tversky loss, and model performance is evaluated using the Dice score. Unet~\cite{ronneberger2015unet} is adopted as specialized task-specific model for lesion segmentation.

\noindent\textbf{3. Super-Resolution} aims to reconstruct high-resolution prostate MRI from low-resolution inputs within the same modality. Specifically, the T2W images from the PROMIS dataset~\cite{ahmed2017diagnostic} are downsampled along the through-plane dimension (apex-base) from 96 to 48 axial slices, forming synthetic low-resolution inputs. The model learns a mapping $f_{SR}: \mathbb{R}^{H\times W\times 48}\rightarrow \mathbb{R}^{H\times W\times 96}$. The network is fine-tuned using L1 reconstruction loss, and evaluated by SSIM between predicted and high-resolution volumes. For super-resolution, Transunet~\cite{chen2021transunet} is selected as specialized task-specific model.

\noindent\textbf{4. Distortion Correction} aims to correct rectal-air-induced geometric distortions in prostate MRI, formulated as
    $f_{cor}: \mathbb{R}^{H\times W\times D}_{dist}\rightarrow \mathbb{R}_{corr}^{H\times W\times D}$. We use the T2W sequence from the few-shot prostate dataset~\cite{li2023prototypical}, which consists of rectal segmentation masks corresponding to each T2W sequence. Distorted inputs are synthetically generated by applying smooth deformation fields derived from rectal masks, where displacement decays gradually from the rectal region to surrounding tissues. Model is fine-tuned using a 330/82/177 (train/validation/test) split. Fine-tuning is guided by L1 reconstruction loss, and the correction quality is evaluated using the structural similarity index (SSIM) between the corrected and reference volumes. We adopt the vit model~\cite{vaswani2017vit} as specialized task-specific model.

\noindent\textbf{5. Modality Translation} learns cross-modal synthesis between prostate MRI contrasts, translating T2W to diffusion-weighted imaging (DWI) within the PROMIS dataset~\cite{ahmed2017diagnostic}. Given paired T2W–DWI volumes, the model performs $f_{trans}: \mathbb{R}^{H\times W\times D}_{T2W}\rightarrow \mathbb{R}^{H\times W\times D}_{DWI}$. We adopt a Denoising Diffusion Probabilistic Model (DDPM)-based strategy~\cite{ho2020ddpm}, where the foundation model predicts noise.
The fine-tuning objective combines L1 reconstruction loss and SSIM regularization, while performance is evaluated using SSIM on the test set. Specialized task-specific model here is DDPM model~\cite{ho2020ddpm}.

\subsection{Task Alignment Measurement}
\label{sec: task align}
To quantify how well each downstream task aligns with a given pretraining objective, we define three complementary representation distance metrics. Each metric measures the divergence between feature embeddings of two models, using the squared maximum mean discrepancy ($\text{MMD}^2$).
Let $T_{down}$ denote a downstream task, and $x\in \mathcal{X}_{down}$ represent an image sampled from the downstream dataset. For a given mode $M$, the encoder-derived feature embedding is denoted as $\phi_M(x)\in\mathbb{R}^d$, where $d$ is the feature dimension. We compute the squared MMD between two feature sets $\Phi_A=\{\phi_A(x_i)\}^N_{i=1}$ and $\Phi_B=\{\phi_B(x_i)\}^N_{i=1}$ as
\begin{equation}
\begin{aligned}
&\text{MMD}^2(\Phi_A,\Phi_B)
=  \frac{1}{N^2}\sum_{i,j} k(\phi_A(x_i), \phi_A(x_j))\\
& + \frac{1}{N^2}\sum_{i,j} k(\phi_B(x_i), \phi_B(x_j)) 
 - \frac{2}{N^2}\sum_{i,j} k(\phi_A(x_i), \phi_B(x_j)),
\end{aligned}
\end{equation}
where $k(\cdot,\cdot)$ is a Gaussian kernel $k(u,v)=exp(-||u-v||^2/\sigma^2)$, and here $\sigma$ is set to 1.

\noindent\textbf{1. Distance to Random-Weight (D2R)} quantifies how far the fine-tuned FM’s representation deviates from that of random-weight FM:
    $\text{D2R}(T_{down}) = \text{MMD}^2(\Phi_{ft},\Phi_{rand}),$  
    where $\Phi_{ft}$ denotes the features from FM fine-tuned on $T_{down}$, and $\Phi_{rand}$ is the features from a model with the same architecture but randomly initialized. A higher D2R indicates that the fine-tuning learns structured, non-random representations that significantly depart from architectural priors.

\noindent\textbf{2. Distance to Pretraining (D2P)} measures the representational shift from the pretrained state to the downstream fine-tuned state:
    $\text{D2P}(T_{down}) = \text{MMD}^2(\Phi_{ft},\Phi_{pre}),$
    where $\Phi_{pre}$ denotes features extracted from the pretrained FM before fine-tuning. A smaller D2P implies a higher degree of alignment to pretraining, indicating that pretrained features remain stable and transferable across tasks.

\noindent\textbf{3. Distance to Task-Specific Training (D2S)} captures the gap between the fine-tuned FM and a model trained from scratch on the same downstream task:
    $\text{D2S}(T_{down}) = \text{MMD}^2(\Phi_{ft},\Phi_{tra}),$
    where $\Phi_{tra}$ denotes features obtained from a model with identical architecture, trained from random initialization solely on $T_{down}$. A smaller D2S suggests that the fine-tuned FM converges toward a similar representational manifold as a trained-from-scrach model.



\begin{figure}[!t]
    \centering
    \includegraphics[width=0.9\linewidth]{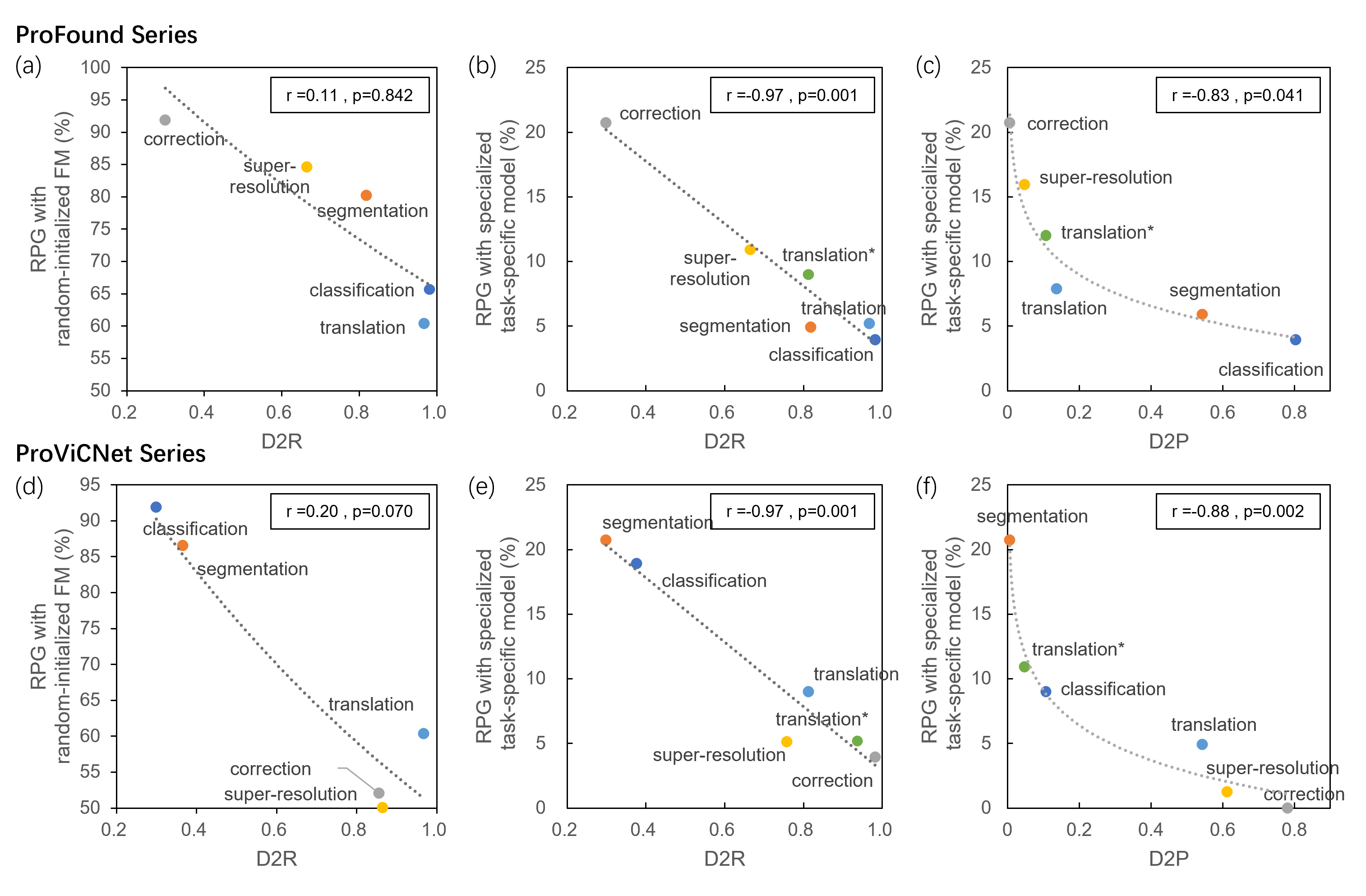}
    \caption{Negative correlation between task alignment (D2R and D2S) and relative performance gain (RPG) across ProFound (a-c) and ProViCNet (d-f) in five downstream tasks. Colored markers denote downstream tasks; asterisks (*) indicate DDPM-based fine-tuning.}
    \label{fig:plots}
    \vspace{-1.5em}
\end{figure}


\subsection{Relative Performance Gain}
\label{sec: rpg}
To measure the effectiveness of foundation model fine-tuning across heterogeneous downstream tasks and evaluation metrics, we define the relative-performance-gain (RPG) as a normalized improvement ratio between the fine-tuned FM and one of the reference models under the varying conditions. 
Formally, for a given downstream task $T_{down}$ and performance metric $P(\cdot)$, the RPG is defined as:
\begin{equation}
\begin{aligned}
&\text{RPG}(T_{down}; M_{ft}, M_{ref}) \\&=
\frac{P(M_{ft}, T_{down}) - P(M_{ref}, T_{down})}
{P(M_{ref}, T_{down})} \times 100\%,
\end{aligned}
\end{equation}
where $M_{ft}$ denotes the fine-tuned FM, and $M_{ref}$ denotes the reference model, the latter of which is specified as either the random-weight FM or the specialized task-specific model in this study. $P(M,T_{down})$ represents the task-specific evaluation metric. Details on the specialized task-specific models and evaluation metrics are provided in Sec.\ref{sec: downstream}.



    

\section{Results and Discussion}
\label{sec:result}

\subsection{Task Alignment Determines the Extent of Gains}


As shown in Fig.\ref{fig:plots} (c,f), the RPG with specialized task-specific model exhibits a strong negative correlation with the representational D2P (Pearson’s r$<$-0.8, p-value$<$0.05). A smaller D2P - indicating higher alignment between the downstream and pretraining objectives - corresponds to larger performance improvements after fine-tuning.
For the MAE-based FM ProFound, tasks that emphasize structural restoration, such as distortion correction and super-resolution, achieve the highest gains (RPG = 20.72\% and 15.91\%), consistent with their low D2P values and reconstruction-driven pretraining objective. In contrast, tasks requiring semantic discrimination, such as PI-RADS classification and lesion segmentation, show larger D2P (0.8034 and 0.5429) and limited RPG (21.23\% and 8.9907\%). For the DINOv2-based FM ProViCNet, which integrates semantic supervision during pretraining, the trend reverses: segmentation and classification tasks yield the smallest D2P (0.0049 and 0.1072) and the largest gains (RPG = 21.23\% and 8.99\%).
As visualized in Fig.\ref{fig:motivation}(b), comparing ProFound fine-tuned from pretraining (fine-tuned FM) with the same architecture trained from scratch (trained-from-scratch FM), points closer to the lower-right corner indicate higher pretraining advantage. For PI-RADS classification task, pretraining brings negligible improvement (sensitivity: 0.6340 vs. 0.6203), consistent with its large D2S (0.3378). These results demonstrate that FM fine-tuning benefits are positively correlated with the alignment between pretraining and downstream objectives.

\begin{figure}[!t]
    \centering
    \includegraphics[width=\linewidth]{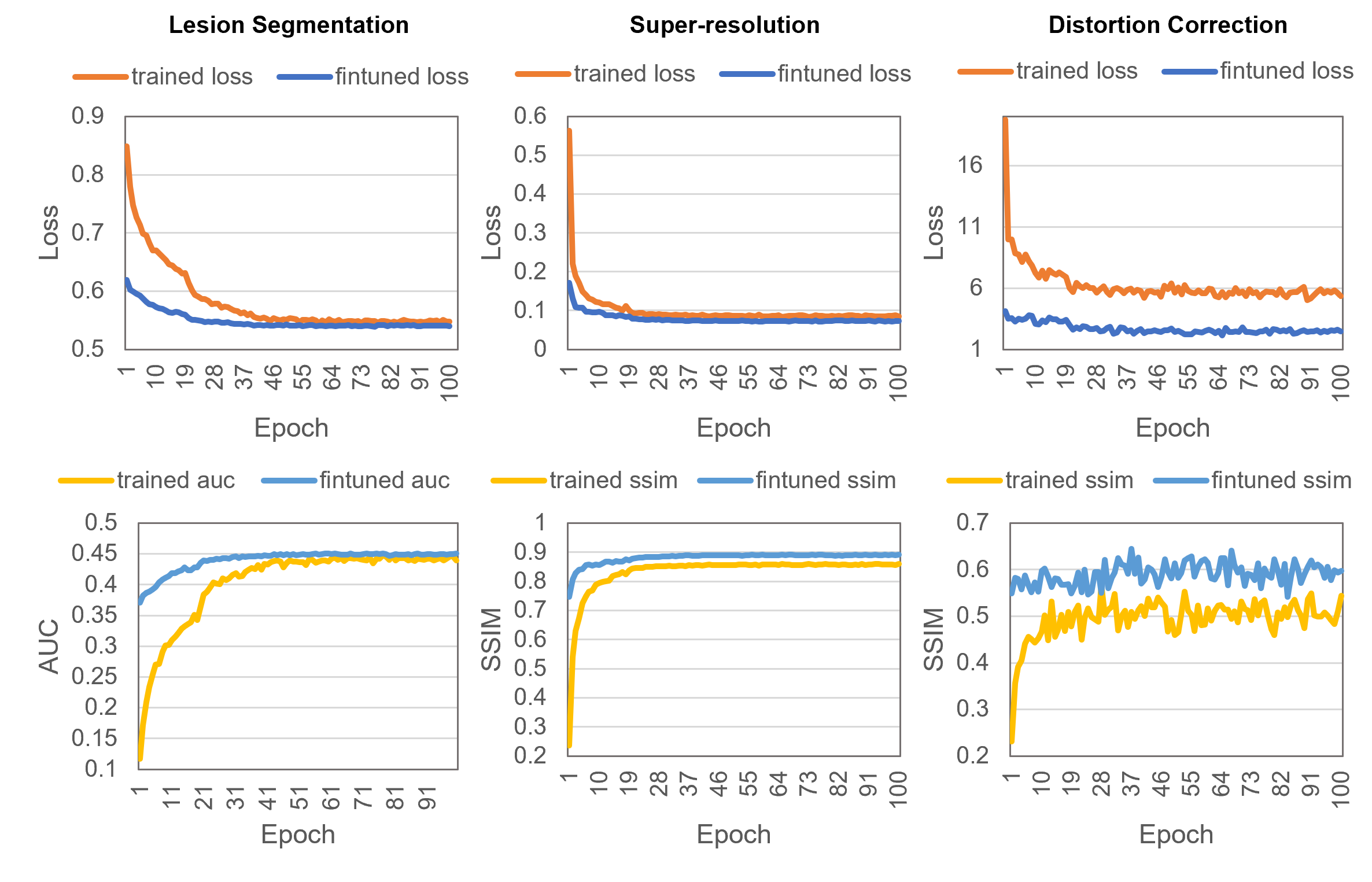}
    \caption{ Training loss (top) and validation metric (bottom) of ProFound over epochs across downstream tasks. 
    }
    \label{fig:loss_curve}
    \vspace{-1.5em}
\end{figure}
\begin{table}
\centering
\scriptsize
\setlength{\tabcolsep}{3pt}
\label{tab:gpu}
\caption{Comparison of GPU-hours for fine-tuning (FT) and training-from-scratch (Scratch) on five downstream tasks.}
\begin{tabular}{l|ccc|ccc} 
\hline
\multicolumn{1}{c|}{\multirow{2}{*}{Tasks}} & \multicolumn{3}{c|}{GPU-hrs ProFound} & \multicolumn{3}{c}{GPU-hrs ProViCNet} \\ 
\cline{2-7}
\multicolumn{1}{c|}{} & FT & Scratch & FT/Scratch & FT & Scratch & FT/Scratch \\ 
\hline
class. & 9.9 & 13.8 & 0.72\% & 8.2 & 18.4 & 0.45\% \\
seg. & 10.4 & 20.7 & 0.50\% & 10.8 & 27.6 & 0.39\% \\
sup.-res. & 8.0 & 19.0 & 0.42\% & 13.7 & 25.3 & 0.54\% \\
dist.corr. & 6.4 & 20.7 & 0.31\% & 13.8 & 27.6 & 0.50\% \\
mod.tra. & 9.4 & 20.7 & 0.45\% & 11.8 & 27.6 & 0.43\% \\
\hline
\end{tabular}
\vspace{-1em}
\end{table}

\subsection{Pretraining Objectives Shape Transfer Preference}

As shown in Fig.\ref{fig:plots} (c,f), the downstream transfer preference of the two FMs exhibits distinct dependencies on their pretraining objectives. For the MAE-based FM (ProFound), the distortion-correction task shows the highest alignment with the pretraining objective, reflected by the lowest representational distance (D2P = 0.0057). In contrast, tasks that require higher-level semantic reasoning, such as classification and segmentation, show the lowest similarity (D2P = 0.8034, 0.5429).  Conversely, for the DINOv2-based FM (ProViCNet), which incorporates semantic supervision during pretraining, the segmentation task achieves the lowest D2P distance (0.0049), while the distortion-correction task shows the highest (0.7811). As illustrated in Fig.\ref{fig:plots} (b,e), even random-weight models exhibit task-dependent embedding biases. After pretraining, task D2P rankings shift - for example, in ProViCNet, the modality-translation task rises from fifth (D2R=0.9368) to second (D2P=0.0472), indicating a clear adaptation of the representation space.

These observations demonstrate that distinct pretraining objectives guide models towards specialized representational structures - MAE emphasizing reconstruction consistency and DINOv2 emphasizing semantic discrimination - which directly determine how feature embeddings align with different downstream task distributions. 

\subsection{Well-Aligned Tasks Adapt Faster}

To assess adaptation dynamics during fine-tuning/training, we tracked loss and performance over epochs for both fine-tuned FM and trained-from-scratch FM (Fig.\ref{fig:loss_curve}). Models initialized from pretraining consistently achieved faster convergence and higher final performance, particularly in tasks exhibiting small D2P to the pretraining distribution. As shown in Tab.\ref{tab:gpu}, GPU-hours for fine-tuned FM and trained-from-scratch FM on five downstream tasks are compared. Referring to Fig.\ref{fig:plots}, tasks with smaller D2P values (e.g., distortion correction task) generally exhibit higher fine-tuning efficiency, requiring less training time relative to scratch training.

\section{Conclusion}
This work highlights that vision foundation model transferability is strongly influenced by the alignment between pretraining objectives and downstream tasks. Strengthening this pretraining-downstream alignment, or alternatively developing approaches that are invariant to misalignment, offers promising directions. Future research may explore adaptive or task-agnostic pretraining strategies to enhance generalization and enable wider, clinically meaningful applications in prostate cancer imaging and beyond.

\section*{Acknowledgments}
\label{sec:acknowledgments}
This work was supported by the International Alliance for Cancer Early Detection, an alliance between Cancer Research UK [EDDAPA-2024/100014, EDDAMC-2021/100011, \\C73666/A31378], Canary Center at Stanford University, the University of Cambridge, OHSU Knight Cancer Institute, University College London and the University of Manchester. This work is supported by the National Institute for Health Research University College London Hospitals Biomedical Research Centre. This research was partly supported by unrestricted philanthropic donations to University College London. 

\printbibliography
\end{document}